\documentclass{ecai} 

\usepackage{latexsym}
\usepackage{amssymb}
\usepackage{amsmath}
\usepackage{amsthm}
\usepackage{booktabs}
\usepackage{enumitem}
\usepackage{graphicx}
\usepackage{color}
\usepackage{algorithm}
\usepackage{algpseudocode}
\usepackage{bm}
\usepackage{multirow}

\newcommand{\BibTeX}{B\kern-.05em{\sc i\kern-.025em b}\kern-.08em\TeX}

\begin{document}


\begin{frontmatter}



\title{Safe Planning and Policy Optimization via World Model Learning}


\author[A,B]{\fnms{Artem}~\snm{Latyshev}\thanks{Corresponding Author. Email: latyshev.ak@phystech.edu.}}
\author[B,C]{\fnms{Gregory}~\snm{Gorbov}}
\author[A,B,C]{\fnms{Aleksandr I.}~\snm{Panov}} 

\address[A]{AIRI, Moscow, Russia}
\address[B]{MIPT, Moscow, Russia}
\address[C]{FRC CSC RAS, Moscow, Russia}


\begin{abstract}
Reinforcement Learning (RL) applications in real-world scenarios must prioritize safety and reliability, which impose strict constraints on agent behavior. Model-based RL leverages predictive world models for action planning and policy optimization, but inherent model inaccuracies can lead to catastrophic failures in safety-critical settings. We propose a novel model-based RL framework that jointly optimizes task performance and safety. To address world model errors, our method incorporates an adaptive mechanism that dynamically switches between model-based planning and direct policy execution. We resolve the objective mismatch problem of traditional model-based approaches using an implicit world model. Furthermore, our framework employs dynamic safety thresholds that adapt to the agent's evolving capabilities, consistently selecting actions that surpass safe policy suggestions in both performance and safety. Experiments demonstrate significant improvements over non-adaptive methods, showing that our approach optimizes safety and performance simultaneously rather than merely meeting minimum safety requirements. The proposed framework achieves robust performance on diverse safety-critical continuous control tasks, outperforming existing methods.
\end{abstract}

\end{frontmatter}


\section{Introduction}

Reinforcement learning (RL) has demonstrated remarkable success in simulated environments through reward maximization \citep{survey_rl}. However, its application to safety-critical domains like autonomous vehicles and robotics necessitates specialized approaches that prioritize safety \citep{SafeControlRlSurvey}. Safe Reinforcement Learning (Safe RL) addresses this need by incorporating cost signals to constrain agent behavior \citep{gu2024reviewsafereinforcementlearning}. Current research predominantly evaluates Safe RL algorithms using the SafetyGymnasium benchmark,\footnote{SafetyGymnasium incorporates all tasks from the original (now unmaintained) SafetyGym benchmark \citep{safetygym}.}, which features complex control problems with Point, Car, Ant, and Doggo agents \citep{ji2023safetygymnasium}. These challenges are typically addressed through either model-free or model-based approaches \cite{omnisafe}.

Model-free methods \citep{gu2024reviewsafereinforcementlearning} suffer from low sample efficiency, often requiring a large number of environment interactions to learn safe behaviors. This limitation is evident in approaches like PPOLagrangian, PPO \citep{schulman2017proximal} combined with PID Lagrangian \citep{stooke2020pidlag}, or CPO \citep{achiam2017cpo}, which may be impractical for real-world applications.

Safe Model-Based RL (Safe MBRL) addresses this issue by integrating transition function learning with safety-aware techniques \citep{SafeControlRlSurvey}. While this paradigm improves sample efficiency through synthetic experience generation \citep{safedreamer,as2022lambda} and model-based planning \citep{hansen2024tdmpc2,liu2020mbrce},\footnote{Sometimes referred to as Background Planning \citep{safedreamer}.} it introduces new challenges, such as: \textbf{uncertainty} in model predictions \citep{SafeControlRlSurvey,moerland2023model,as2022lambda}, \textbf{objective mismatch} problems between model optimization and task performance \citep{moerland2023model,wei2023objmismatch}.

To mitigate world model uncertainty, the agent must perform efficient safe exploration. However, safe exploration presents a fundamental conflict: \textit{learning an accurate world model requires encountering potentially unsafe states, while perfect constraint avoidance demands complete prior knowledge of hazardous situations}. The Constrained Markov Decision Process (CMDP) framework \citep{altman2021constrained} addresses this by permitting controlled safety violations during training through inequality constraints, though this inherently limits final policy safety.

This \textbf{safety-performance trade-off} becomes particularly acute in near-zero-violation scenarios \citep{safedreamer}, where selecting appropriate safety thresholds requires a fine balance: \textit{the threshold must be large enough to allow sufficient exploration yet small enough to ensure high safety in the final policy}.

\textit{Epistemic uncertainty} further complicates this situation. This uncertainty arises from insufficient knowledge about the environment. It becomes particularly crucial when using synthetic data generated by world models, as inaccuracies can propagate and compromise safety. Several approaches address this challenge: \citet{as2022lambda} employs a Bayesian framework to learn safe policies while accounting for model uncertainty; \citet{liu2020mbrce} utilizes ensemble methods for robust planning.

\textbf{However, these methods neglect the agent's evolving capabilities, potentially wasting resources on precise world modeling when direct policy learning might suffice.} This presents an important trade-off in sample efficiency.

Safe reinforcement learning in high-dimensional continuous control tasks requires balancing task performance with stringent safety constraints. While model-based RL offers sample efficiency, it faces challenges in safety-critical settings: inaccurate world models may lead to catastrophic failures, and static safety thresholds struggle to adapt to evolving agent capabilities. We propose \textbf{S}afe \textbf{P}lanning and p\textbf{O}licy \textbf{O}ptimization via \textbf{W}orld model \textbf{L}earning (SPOWL), a novel model-based RL framework that unifies safe policy optimization with adaptive model-based planning to address these limitations.

\textbf{Contributions:}
\begin{itemize}
    \item We empirically demonstrate that SPOWL robustly achieves near-zero safety violations across diverse continuous control tasks, outperforming baselines while maintaining high task performance.
    \item We introduce a hybrid framework that dynamically switches between a safe policy (trained via constrained optimization) and model-based planning, mitigating world model inaccuracies.
    \item We identify the limitation of fixed safety thresholds in planning and propose a cost-value-function-driven mechanism to adjust thresholds as the agent learns, enabling progressive risk-taking.
    \item We address the objective mismatch problem in Safe MBRL by adopting implicit world models, where planning occurs directly in latent space without decoding, aligning planning and policy objectives.
\end{itemize}


\section{Related Work}

Safe reinforcement learning (RL) methods can be broadly categorized into \textbf{model-free} and \textbf{model-based} approaches. Model-free methods typically rely on Lagrangian multipliers derived from the Constrained Markov Decision Process (CMDP) dual optimization problem, while model-based approaches use world models or planning techniques to supplement real-world interactions with simulated experience.

\subsection{Model-Free Approaches}
Lagrangian methods dominate model-free safe RL, spanning both off-policy and on-policy algorithms. The SAC-Lagrangian method \cite{sac_l,stooke2020pidlag} integrates the entropy-driven exploration of Soft Actor-Critic \citep{sac} with Lagrangian optimization \cite{stooke2020pidlag}, achieving better sample efficiency than early approaches like Constrained Policy Optimization (CPO) \cite{achiam2017cpo}. However, like all off-policy methods, SAC-Lagrangian suffers from value estimation bias and lacks formal safety guarantees.  

On-policy alternatives such as PPO-Lagrangian \cite{stooke2020pidlag} offer more stable policy updates at the cost of higher sample complexity. Theoretical safety guarantees remain challenging due to the inherent complexity of RL dynamics and high-dimensional state-action spaces. While methods like RESPO \cite{respo} provide formal guarantees, they often rely on restrictive assumptions. As an example, they require an initially feasible policy, LBSGD optimization \cite{lbsgd}.  

Variants of Lagrangian methods aim to improve constraint satisfaction. For example, the Augmented Lagrangian approach \cite{auglag,as2022lambda,safedreamer} introduces additional penalty terms to push solutions toward feasible regions more aggressively. PID Lagrangian uses PID control to compute lagrange multipliers that solves the problem of oscillation near safety threshold \citep{stooke2020pidlag}.

\subsection{Model-Based Approaches}
Model-based safe RL methods generally outperform model-free counterparts in sample efficiency \cite{ji2024omnisafe}. MBPPOL \cite{mbppol} trains a PPO-Lagrangian agent using synthetic data generated by an ensemble of world models to address epistemic uncertainty. However, this method does not employ planning, limiting its utility to passive data generation. Additionally, its reliance on manually generated LiDAR data reduces generalizability to other benchmarks.  

In MBRCE \cite{mbrce}, the authors combine an ensemble world model, a LightGBM-based cost predictor, and a robust cross-entropy (RCE) planner. While the ensemble mitigates model inaccuracies, the framework lacks memory mechanisms---a key feature of modern approaches like Dreamer \cite{dreamerv3,safedreamer}---making it ineffective for memory-dependent tasks.  

The SafeDreamer \cite{safedreamer} achieves state-of-the-art performance on the SafetyGym benchmark \cite{ji2023safetygymnasium} (tested in image-based tasks). It extends the Dreamer framework \citep{dreamerv3} with a constrained cross-entropy planning (CCEM; \citep{wen2018cce}), and a safe background policy for trajectory generation \citep{auglag}. Despite its advancements, Dreamer-based methods struggle with large continuous spaces (see TD-MPC2 comparisons \citep{hansen2024tdmpc2}).


\section{Background}
\label{sec:background}

Effective decision-making in safety-critical scenarios requires balancing performance with constraint satisfaction. These requirements are formally captured by Constrained Markov Decision Processes, which extend traditional RL frameworks with safety constraints. This work integrates RL and planning through Model-Based RL (MBRL), leveraging the strengths of CMDPs for formal safety guarantees and Model Predictive Control (MPC) for real-time adaptability.

\subsection{Constrained Markov Decision Processes}
The standard framework for RL is the Markov Decision Process (MDP) \citep{puterman2014markov}, which models an agent interacting with an environment through states, actions, and rewards. For safety-critical tasks, Constrained MDPs (CMDPs) \citep{altman2021constrained} extend MDPs by introducing constraints on cumulative costs, ensuring policies satisfy predefined safety limits.

A CMDP is defined by the tuple $\langle \mathcal{S}, \mathcal{A}, p, R, C, d, \rho, \gamma, \gamma_c \rangle$, where $\mathcal{S}$ and $\mathcal{A}$ represent continuous state and action spaces. The transition dynamics function $p(s' | s, a)$ specifies the probability density of transitioning to state $s' \in \mathcal{S}$ when taking action $a \in \mathcal{A}$ from state $s \in \mathcal{S}$, while $\rho(s_0)$ defines the initial state distribution. The reward function $R: \mathcal{S} \times \mathcal{A} \to \mathbb{R}$ determines the immediate reward $R(s, a)$ for each state-action pair, and the cost function $C: \mathcal{S} \times \mathcal{A} \to \mathbb{R}_{\geq 0}$ assigns non-negative penalties for safety violations. The safety constraints are enforced through the cost threshold $ d \in \mathbb{R}_{\geq 0}$, which bounds the expected cumulative cost, with $\gamma, \gamma_c \in [0,1)$ serving as discount factors for rewards and costs, respectively.

The agent's objective is to learn a policy $\pi: \mathcal{S} \to \mathcal{P}(\mathcal{A})$ (a probability distribution over actions for every state) that maximizes the expected cumulative discounted reward:

\begin{equation}
J(\pi) = \mathbb{E}_{\pi} \left[ \sum_{t=0}^{\infty} \gamma^t R(s_t, a_t) \right],
\end{equation}
while satisfying the safety constraint on expected cumulative discounted cost:

\begin{equation}
J_c(\pi) = \mathbb{E}_{\pi} \left[ \sum_{t=0}^{\infty} \gamma_c^t C(s_t, a_t) \right] \leq d.
\end{equation}

\subsection{Model-Based Reinforcement Learning}

MBRL enhances efficiency and safety by learning an approximate world model of the dynamics $p(s' | s, a)$ and cost/reward functions from data \citep{moerland2023model}. This learned world model facilitates the formation of a meaningful latent state space. Access to transition dynamics enables planning, which is then leveraged for control. 

We consider Model Predictive Control (MPC) \cite{negenborn2005mpclearning,hewing2020safempclearning} as a widely used approach that optimizes action sequences over a finite horizon $H$:
\begin{equation}
\label{eq:mpc}
    \pi(s_t) = \text{argmax}_{a_{t:t+H}} J^M(a_{t:t+H}),
\end{equation}
where $J^M$ estimates cumulative rewards based on world model predictions. The agent executes the first action of the sequence and replans at each subsequent step.

To enforce safety, we constrain MPC to feasible action sequences \citep{safedreamer,wen2018cce}:
\begin{multline}
    \pi(s_t) = \text{argmax}_{a_{t:t+H} \in A_s} J^M(a_{t:t+H}), \\ \text{s.t. } A_s = \{a_{t:t+H}| J^M_c(a_{t:t+H}) < d_M\},
\end{multline}
where $J^M_c$ estimates cumulative costs and $d_M$ is a safety threshold analogous to $d$ in CMDPs.


\section{SPOWL: Adaptive Planning with Safe Policy}

SPOWL is a safe model-based reinforcement learning framework designed for continuous control tasks with safety constraints. It simultaneously maximizes task performance while minimizing constraint violations through four core components.

\textbf{World Model}, a predictive model, learns latent representations, reward and cost functions, and corresponding value functions to enable safe planning. It is based on TD-MPC2 \citep{hansen2024tdmpc2}, a robust implicit model-predictive control algorithm for continuous MDPs.

\textbf{Safe Policy} provides a risk-averse action prior for decision-making when the world model generates unreliable plans. We employ an Augmented Lagrangian method \citep{auglag,as2022lambda,safedreamer} for policy optimization.

\textbf{Safe Improvement Planning} searches for trajectories that outperform the safe policy in both performance and safety. This is our proposed mechanism---an alternative to Constrained Cross-Entropy (CCE) \citep{wen2018cce}---that dynamically adjusts safety thresholds based on the cost value function recommendations, unlike CCE's fixed thresholds.

\textbf{Adaptive Decision Making} module evaluates the value functions of proposed actions (from either the world model or policy) to guide switching between planning and policy execution.

\subsection{World Model}
\label{subsec:worldmodel}

\begin{figure}[h]
\centering
\includegraphics[width=\columnwidth]{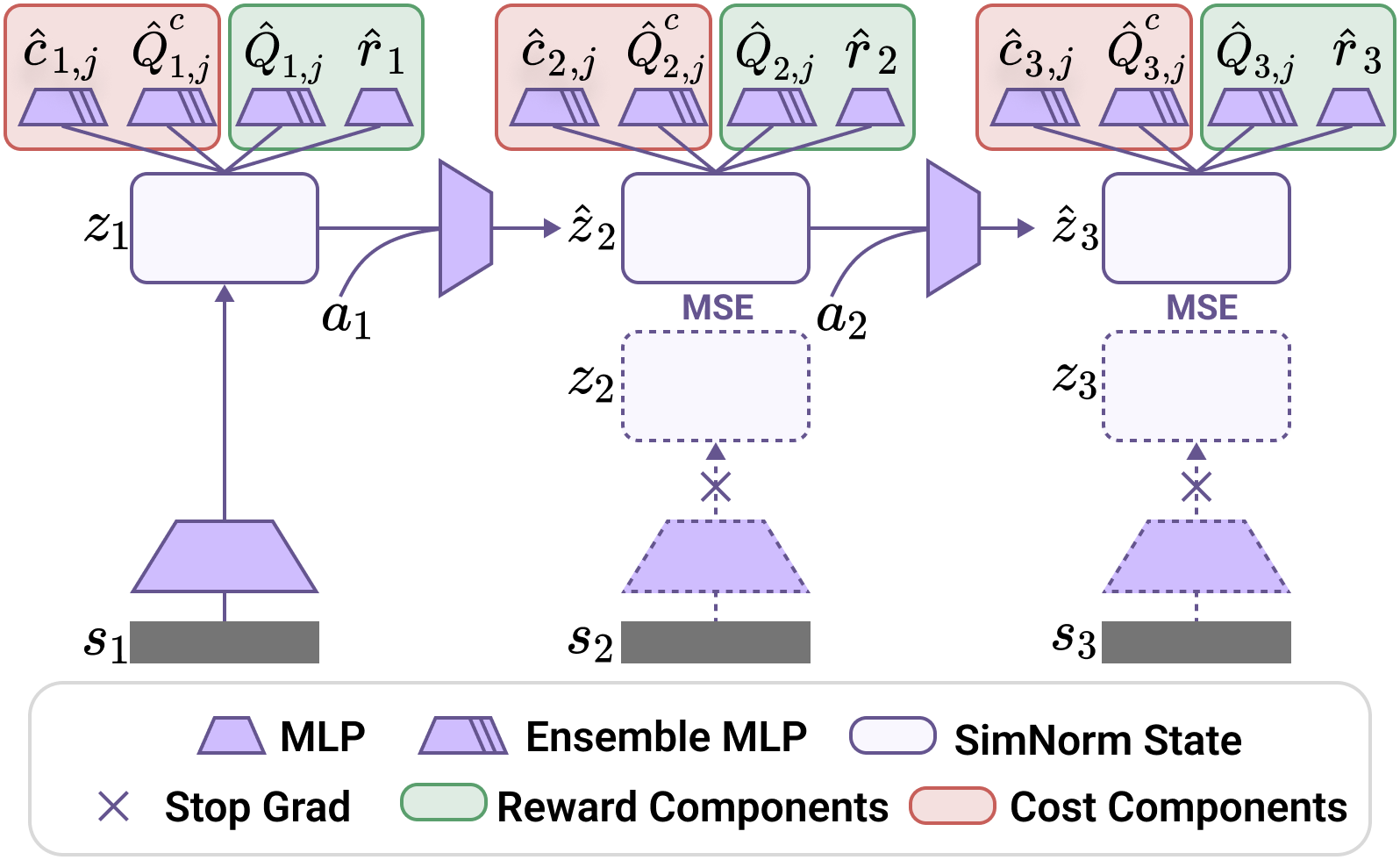}
\caption{\textbf{World model architecture.} Neural networks implement all components, with the latent state space following the SimNorm scheme \cite{hansen2024tdmpc2}. Cost and reward components use discrete regression optimization \cite{hansen2024tdmpc2}. Dashed components generate targets for latent states to provide consistency.}
\label{fig:architecture}
\end{figure}

The world model in SPOWL is value-equivalent, belonging to a subclass of implicit world models \cite{moerland2023model}. This design aligns the model's objective with the agent's, eliminating the need for observation decoding. The model simultaneously approximates both cost and reward value functions while maintaining consistent latent state representations. As shown in Figure~\ref{fig:architecture}, each component is implemented as either a Multi-Layer Perceptron (MLP) or an MLP ensemble:

\begin{equation*}
    \begin{array}{ll}
        z_t = h(s_t) & \text{Encoder} \\
        \hat{z}_{t+1} = f(z_t, a_t) & \text{Dynamics} \\
        \hat{r}_t = \hat{R}(z_t, a_t) & \text{Reward model} \\
        \hat{Q}_{t,j} = \hat{Q}_j(z_t, a_t) & \text{Value ensemble} \\
        \hat{c}_t = \hat{C}(z_t, a_t) & \text{Cost model ensemble} \\
        \hat{Q}^c_{t,j} = \hat{Q}^c_j(z_t, a_t) & \text{Cost value ensemble},
    \end{array}
\end{equation*}
where $s_t$ represents the environment state and $a_t$ denotes the agent's action at time step $t$.

The world model employs the SimNorm scheme \citep{hansen2024tdmpc2} to normalize states and align encoder outputs $z$ with dynamics predictions $\hat{z}$ in a shared latent space. The complete loss function incorporates this alignment constraint along with temporal difference (TD) learning: $\mathcal{L}\left(\theta\right) \doteq \mathop{\mathbb{E}}_{\left(s, a, r, c, s'\right)_{0:H} \sim \mathcal{B}} \sum_{t=0}^{H} \lambda^{t} L_t,$
where
\begin{multline}
     L_t = \underbrace{\|\ \hat{z}_{t}' - \operatorname{sg}(z_t') \|^{2}_{2}}_{\text{Consistency loss}} + \underbrace{\operatorname{CE}(\hat{r}_{t}, r_{t})}_{\text{Reward loss}} + \underbrace{\frac{1}{N}\sum_j^N\operatorname{CE}(\hat{Q}_{t,j}, Q_{t})}_{\text{Value loss}} +\\
    \underbrace{\frac{1}{N}\sum_j^N\operatorname{CE}(\hat{c}_{t,j}, c_{t})}_{\text{Cost loss}}+ \underbrace{\frac{1}{N}\sum_j^N\operatorname{CE}(\hat{Q}^c_{t,j}, Q^c_{t})}_{\text{Cost Value loss}},
\end{multline}
here $\operatorname{sg}(\cdot)$ denotes the stop-gradient operator, $\lambda \in (0,1]$ controls the weighting of future prediction errors, and $\operatorname{CE}$ represents the cross-entropy loss. 

We employ discrete regression in $\log$-transformed space to optimize cost, reward, and value functions by minimizing cross-entropy. This approach proves particularly effective for tasks with wide reward magnitude variations \cite{dreamerv3,hansen2024tdmpc2}. The TD targets for these functions at time step $t$ are defined as:
\begin{equation}
\label{eq:qtargets}
    \begin{array}{ll}
        Q_{t} \doteq r_{t} + \gamma \overline{Q}(\hat{z}_{t}', \hat{a}_{t}') & Q^c_{t} \doteq c_{t} + \gamma_c \overline{Q}^c(\hat{z}_{t}', \hat{a}_{t}'),\\
    \end{array}
\end{equation}
where $\overline{Q}$ and $\overline{Q}^c$ represent exponential moving averages of $Q$ and $Q^c$ respectively, following standard reinforcement learning practice \cite{lillicrap2015continuous}. The actions $\hat{a}_{t}'$ are determined by a policy learned through a separate loss function (see Section~\ref{sec:pilearn}).

For value ensemble aggregation, we compute reward values as the minimum of two randomly subsampled $Q_j$ estimates from five ensemble heads \citep{hansen2024tdmpc2,chen2021randomized} (this helps to reduce bias in value estimation), while cost values use the ensemble average of all $Q_j^c$ functions (see Section~\ref{subsec:ablate} for detailed discussion).

\subsection{Safe Policy Learning with Lagrangian}
\label{sec:pilearn}

In SPOWL, we combine two powerful approaches for policy optimization: \textit{the stochastic maximum entropy framework} \cite{sac} for effective exploration, and \textit{the Augmented Lagrangian method} \cite{auglag,as2022lambda} for constrained optimization. This hybrid approach leverages the exploration benefits of entropy maximization while maintaining rigorous safety constraints through a dual-penalty mechanism.

The policy objective simultaneously maximizes expected reward returns and minimizes safety violations:
\begin{equation}
\label{eq:policy-objective}
    \mathcal{L}_{\pi}(\theta) \doteq \mathop{\mathbb{E}}_{(s,a)_{0:H} \sim \mathcal{B}} \sum_{t=0}^{H} \lambda^{t}L_{\pi}^t + \Psi\left(\Delta, \lambda_l^k, \mu^k\right),
\end{equation}
where the reward component follows the maximum entropy RL formulation \citep{sac}:
\begin{equation}
    L_{\pi}^t = -\alpha Q(z_{t}, \pi(z_{t})) - \beta \mathcal{H}(\pi(\cdot | \hat{z}_{t})),
\end{equation}
and the latent states evolve through forward dynamics: 
\begin{equation}
    \hat{z}_{t+1} = f(\hat{z}_{t}, a_{t}), \quad \text{initialized as } \hat{z}_{0} = h(s_{0}).
\end{equation}

The safety aware term $\Psi$ implements Augmented Lagrangian scheme with three key components:
\begin{itemize}
    \item Lagrangian multiplier $\lambda_l^k$ for constraint enforcement,
    \item Non-decreasing penalty $\mu^{k} = \max(\mu^{k-1}(\nu + 1.0), 1.0)$,
    \item Constraint violation measure $\Delta = \mathop{\mathbb{E}}_{(s,a)_{0:H} \sim \mathcal{B}} \left(Q^c(\hat{z}_t)-b\right)$,
\end{itemize}
where $b=0.1$ defines our strict safety threshold (enforcing near-zero cost operations), $\nu > 0$ controls the penalty growth rate, and $k$ indexes optimization steps. The multipliers update according to:
\begin{equation}
\label{eq:lag}
\Psi,\lambda_l^{k+1} =
    \begin{cases}
    \lambda_l^k\Delta+\frac{\mu^k}{2}\Delta^2, \lambda_l^k+\mu^k\Delta & \text {if } \lambda_l^k+\mu^k\Delta \geq 0, \\
    -\frac{\left(\lambda_l^k\right)^2}{2 \mu^k}, 0 & \text {otherwise.}
    \end{cases}
\end{equation}

\subsection{Safe Improvement Planning}
\label{sec:safeplan}

Constrained Cross-Entropy (CCE) is a planning method for CMDPs \citep{wen2018cce}. It adapts the standard cross-entropy method: iterative selection of ``elite'' trajectories (top-$k$ based on reward value $J$) producing a local plan with some horizon $H$. CCE imposes constraints during this search: each trajectory $\tau$ is characterized by a cost value $J_c(\tau)$, and only feasible trajectories, where $J_c(\tau) < d^c$, are added to the elite set. However, CCE assumes perfect true estimates, ignoring approximation errors in $J_c$, the world model $M$, and the reward value $J$.

\begin{figure}[h]
    \centering
    \includegraphics[width=\columnwidth]{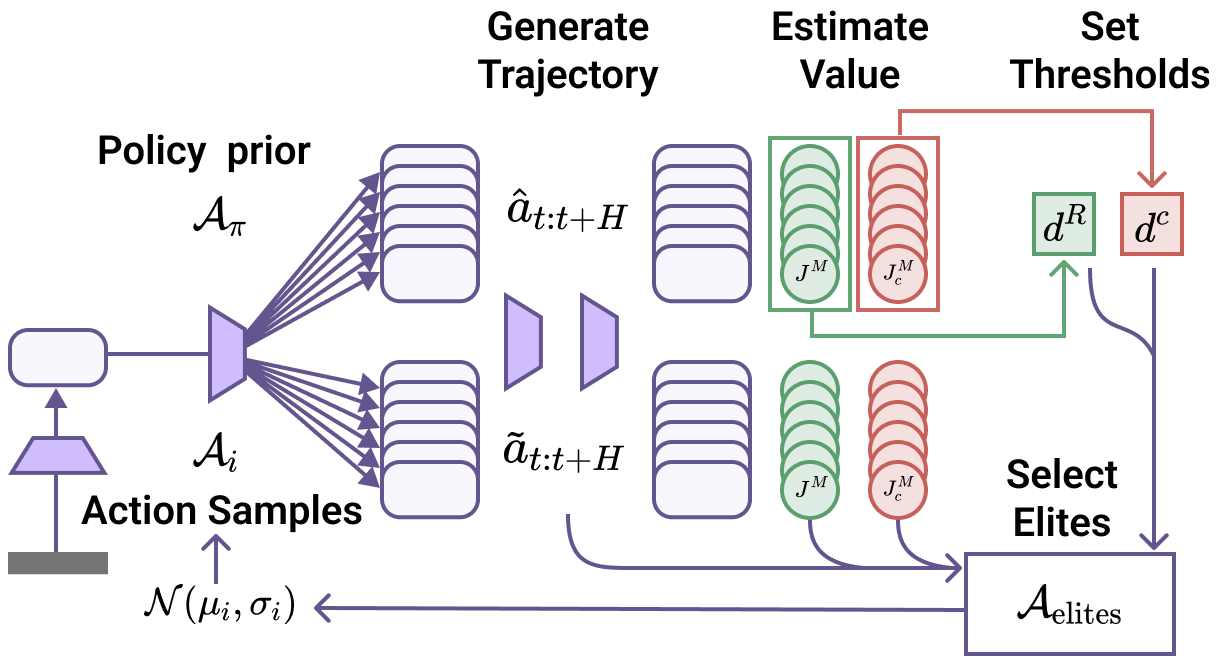}
    \caption{\textbf{SPOWL Planning Process.} The agent evaluates imagined trajectories from: policy-generated sequences (Policy Prior), and samples from a normal distribution parameterized by statistics from previous iterations (Action Samples). Selection thresholds $d^R$ and $d^c$ are derived from the Policy Prior set.}
    \label{fig:planning}
\end{figure}

In SPOWL, we follow a similar elite trajectory selection paradigm but dynamically adjust the safety threshold based on current cost estimates for policy-generated trajectories. The raw state $s_t$ is first encoded into a latent representation $z_t$. Then the local plan is constructed in several steps (see Figure~\ref{fig:planning}):

\paragraph{Action Sequence Generation.}
Using the world model $M$, we generate and evaluate trajectories $\tau$ with cost and value estimates:
\begin{equation}
\label{eq:wmestimates}
    \begin{array}{l}
        \tau=(z_t, a_t, \ldots, z_{t+H}, a_{t+H}), \\
        z_{t+1} = f(z_t, a_t), \quad a_t \sim \mathcal{N}(\mu_a, \sigma_a) \operatorname{or} \pi(z_t), \\
        J^M= \sum_{i=0}^{H-1} \gamma^i \hat{r}_{t+i} + \gamma^H \hat{Q}_\text{avg}(z_{t+H}, a_{t+H}),\\
        J^M_c=\sum_{i=0}^{H-1} \gamma_c^i \hat{c}_{\text{max}, t+i} + \gamma_c^H \hat{Q}_{c,\text{avg}}(z_{t+H}, a_{t+H}),\\
    \end{array}
\end{equation}
where the subscript ``avg'' denotes ensemble averaging. We generate two action sequence sets: \textit{Policy Prior} $\mathcal{A}_\pi = \{\hat{a}_{t:t+H}\}$, actions sampled from policy $\pi$, and \textit{Action Samples} $\mathcal{A}_i = \{\tilde{a}_{t:t+H}\}$, actions drawn from $\mathcal{N}(\mu_i, \sigma_i)$, where $\mu_i,\sigma_i$ are derived from the previous $i-1$ iteration's elite set.

\paragraph{Selection Phase.}
We select improvement trajectories $\mathcal{A}_\text{impr}$ using thresholds computed from the Policy Prior:
\begin{equation}
\label{eq:thresholds}
    \begin{array}{l}
    d^R = \frac{1}{|\{\hat{a}_{t:t+H}\}|} \sum_{\mathbf{a} \in \{\hat{a}_{t:t+H}\}} J^M(\mathbf{a}), \\
    d^c = \frac{1}{|\{\hat{a}_{t:t+H}\}|} \sum_{\mathbf{a} \in \{\hat{a}_{t:t+H}\}} J^M_c(\mathbf{a}), \\
    \end{array}
\end{equation}
with selection criteria $J^M \geq d^R$ and $J^M_c \leq d^c$ (see Supplementary A, Algorithm~1). These ensure trajectories are both high-reward and low-cost.

If no trajectories meet these criteria, we default to $\mathcal{A}_\pi$. When more than $k$ improvement trajectories exist, we select only the top-$k$ by $J^M$ (see Supplementary A, Algorithm~1).

\subsection{Adaptive Decision Making}
\label{sec:adm}
SPOWL employs a dual-action selection mechanism, combining two approaches.
\begin{itemize}
    \item A \textbf{safe global policy} $\pi$ provides stable, reliable actions unaffected by world model prediction uncertainties (Section~\ref{sec:pilearn}).
    \item A \textbf{local planning} approach offers potentially higher-value and safer actions using current world model estimates (Section~\ref{sec:safeplan}).
\end{itemize} 

The policy action represents a conservative baseline, while the local plan action may yield superior performance by leveraging up-to-date model information. The action selection is governed by value function comparisons. If $\hat{Q}_{\text{avg}}(z_t,a_{\text{plan}}) \geq \hat{Q}_{\text{avg}}(z_t,\pi(z_t))$ and $\hat{Q}_{c,\text{avg}}(z_t,a_{\text{plan}}) \leq \hat{Q}_{c,\text{avg}}(z_t,\pi(z_t))$, then $a_\text{env} = a_\text{plan}$. Otherwise,  $a_\text{env} = \pi(z_t)$. This decision process is formalized in Supplementary A, Algorithm~2.


\section{Experiments}

We conduct a series of experiments to demonstrate the challenges discussed earlier and show how SPOWL and its components address these issues. For evaluation, we use SafetyGymnasium~\citep{ji2023safetygymnasium}---a unified framework for testing SafeRL algorithms. This benchmark includes diverse agent embodiments and various continuous control tasks with differing levels of difficulty, enabling comprehensive testing of any SafeRL algorithm.

We employ three principal metrics to assess algorithm performance.
\begin{itemize}
    \item \textbf{Return} (episode reward): the total accumulated reward $\sum_{i=0}^{H} r_i$ during an evaluation episode, measuring task performance (higher values preferred).
    \item \textbf{Costs} (episode cost): the total accumulated cost $\sum_{i=0}^{H} c_i$ during an evaluation episode, quantifying safety violations (lower values preferred). 
    \item \textbf{Cost Rate}: the average cost per timestep $\frac{1}{t}\sum_{\tau=0}^{t} c_\tau$ across training, indicating safety during learning (lower values preferred).
\end{itemize} 

Our experimental setup is designed to answer the following key questions:

\textbf{Adaptive Threshold:} Does SPOWL's adaptive safety threshold enhance final performance compared to fixed thresholds?

\textbf{Dynamic Switching:} Can strategic switching between world model plans and global policy effectively balance their respective strengths?

\textbf{Ablation Studies:} How do SPOWL's hyperparameters affect agent safety and performance?

\textbf{Comparative Analysis:} How does SPOWL compare to existing SafeRL methods?

\subsection{Adaptive Threshold}

We implement the traditional Constrained Cross-Entropy (CCE) scheme for planning with a world model \citep{safedreamer,wen2018cce}.\footnote{The world model used here is the same as described in Section~\ref{subsec:worldmodel}.} This approach requires setting a fixed threshold $d_\text{plan}$ to identify feasible trajectories---those satisfying $J_c^M < d_\text{plan}$.

We consider two approaches for evaluating $J_c^M$: \textit{Global value estimation} computed using Equation~(\ref{eq:wmestimates}), providing long-term cost predictions, \textit{Local value estimation} uses only the sum of immediate cost predictions over a finite horizon $H$, excluding $\hat{Q}_{c,\text{avg}}$.

\textbf{The global estimation} method provides more accurate long-term predictions and enables greater safety guarantees. However, this approach has two significant limitations: the value function evolves concurrently with policy updates during training, and temporal difference (TD) learning may introduce estimation bias. These inherent features create substantial challenges in establishing appropriate safety thresholds that remain valid throughout the learning process.

\begin{figure}[h]
    \centering
    \includegraphics[width=0.8\columnwidth]{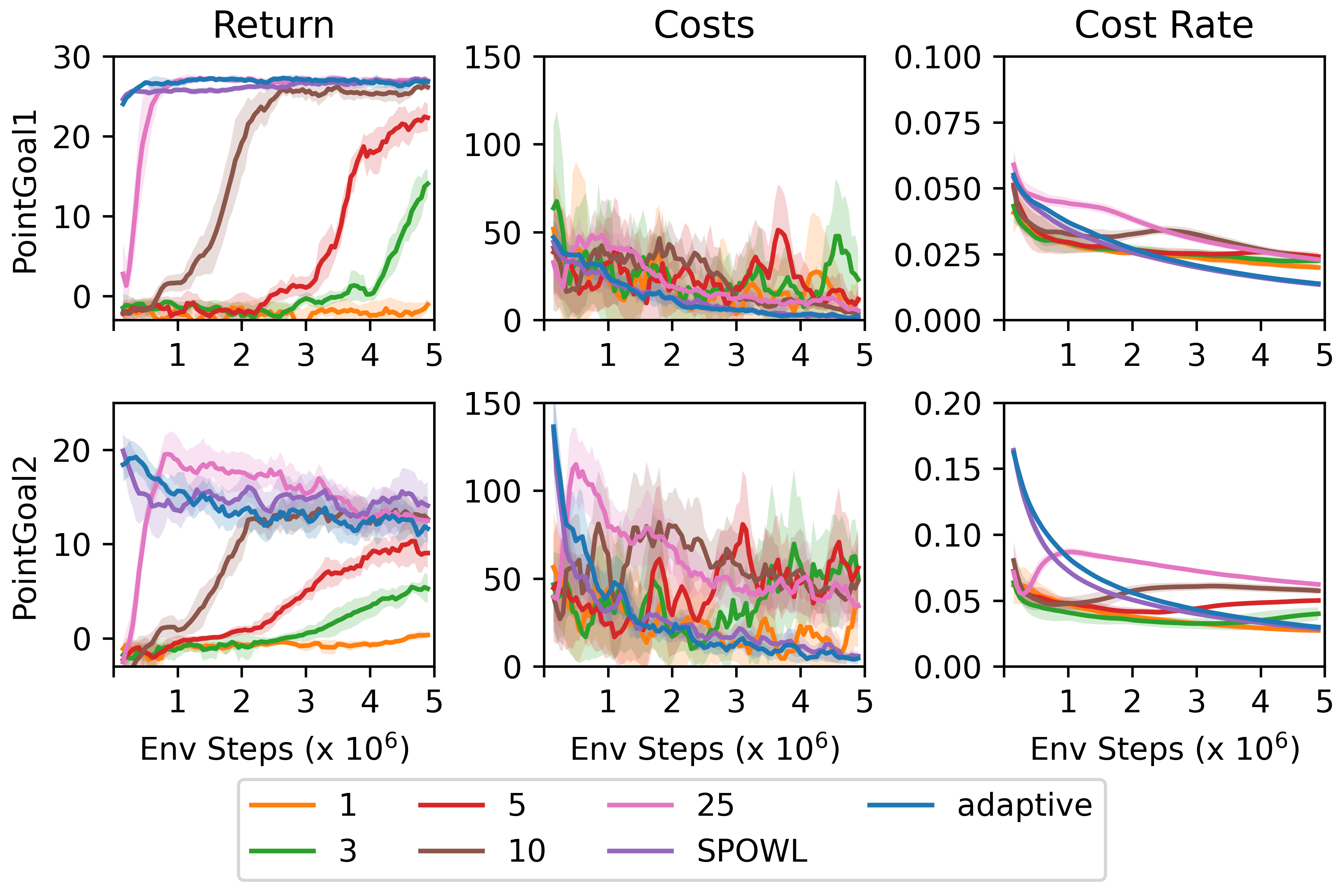}
    \caption{\textbf{Global estimation} for ablation safety thresholds ($d_\text{plan}=1, 3, 5, 10, 25$) of Constrained Cross Entropy planning. SPOWL denotes agent with all components, ``adaptive'' uses only Adaptive Threshold without Dynamic Policy Switching. }
    \label{fig:cceg}
\end{figure}

As shown in Figure~\ref{fig:cceg}, we observe a critical trade-off in threshold selection. Excessively low thresholds restrict environmental exploration, resulting in stagnant learning and negligible performance gains. Conversely, increasing the threshold leads to an exponential rise in safety violations.

\begin{figure}[h]
    \centering
    \includegraphics[width=0.8\columnwidth]{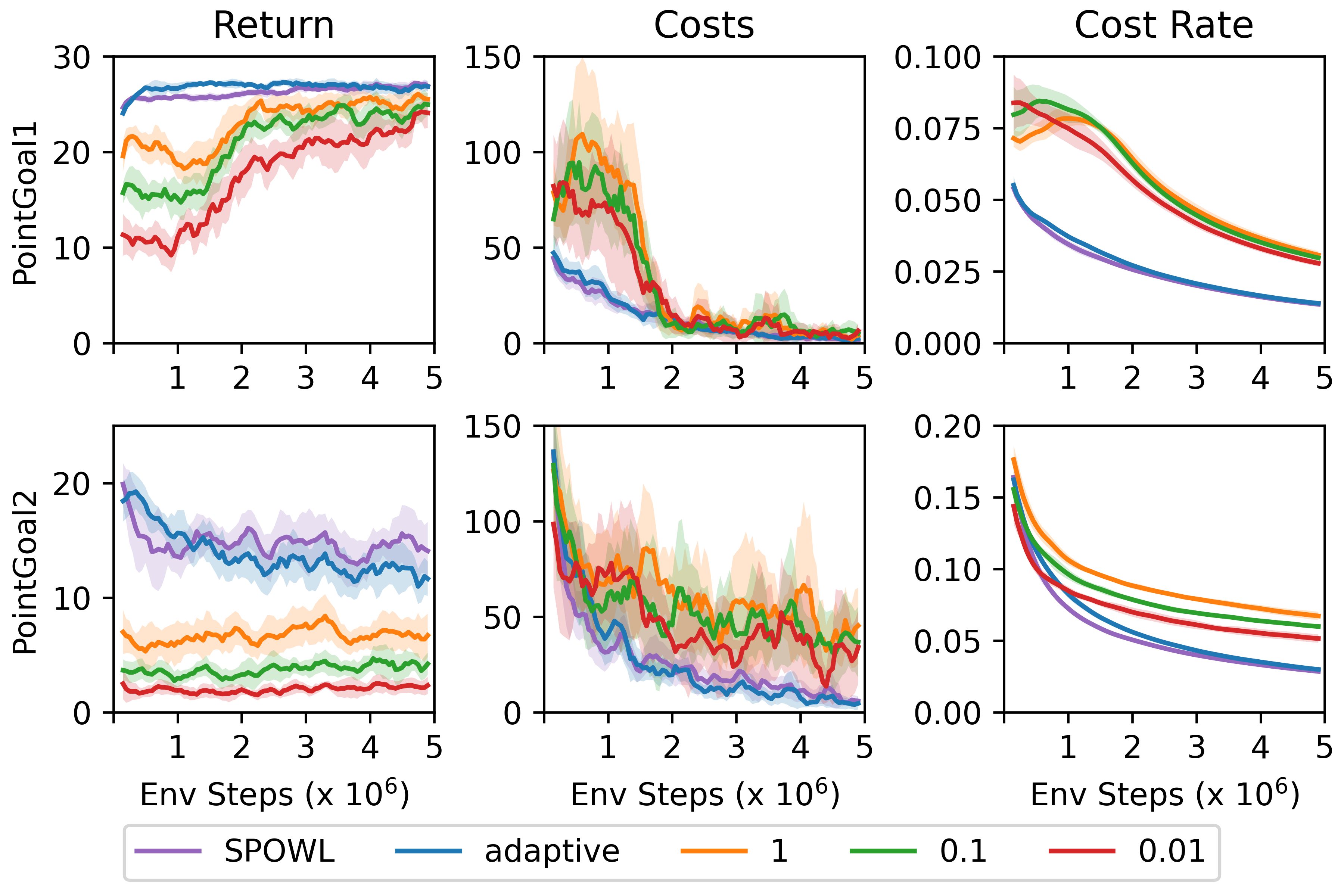}
    \caption{\textbf{Local estimation} for ablation safety thresholds ($d_\text{plan}=0.01, 0.1, 1$) of Constrained Cross Entropy planning. SPOWL denotes agent with all components, ``adaptive'' uses only Adaptive Threshold without Dynamic Policy Switching. }
    \label{fig:ccel}
\end{figure}

\textbf{The local estimation} method suffers from limited future visibility. In environments with sparse cost signals, this approach may detect cost violations too late. Our experiments demonstrate that local estimation requires extremely small thresholds\footnote{For horizon $H=3$, cost violations must not exceed $d_\text{plan}$. With $d_\text{plan}=1$, an episode of length 200 could accumulate approximately 70 violations.} to maintain safety, but this significantly compromises reward performance (see Figure~\ref{fig:ccel}).

\begin{figure}[h]
    \centering
    \includegraphics[width=\columnwidth]{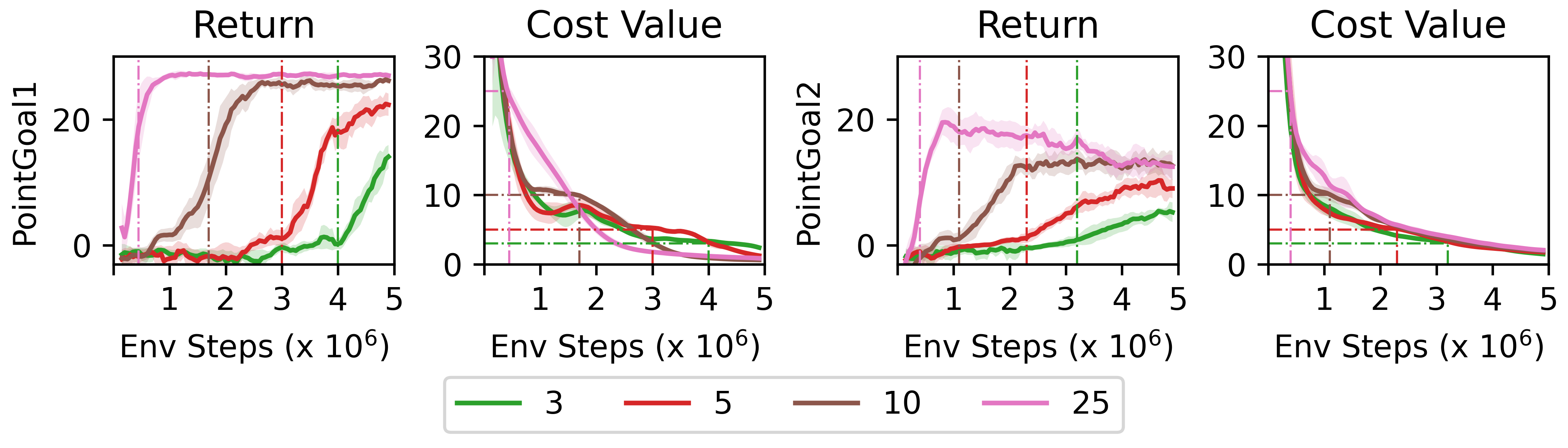}
    \caption{The performance of the Constrained Cross Entropy with global estimation starts to rise then the estimation falls below the threshold (dashed lines). The mean cost value estimation is high for early steps.}
    \label{fig:thr}
\end{figure}

The comparison reveals fundamental differences between the approaches:
\begin{itemize}
    \item \textbf{Local estimation} achieves faster reward convergence due to optimistic early-stage cost predictions (resulting from data imbalance) and sparse cost signals, facilitating exploration.
    \item \textbf{Global estimation} initially propose high cost values (see Figure~\ref{fig:thr}) because the policy is unsafe, severely limiting exploration. Performance improves only when cost estimates fall below the threshold (Figure~\ref{fig:thr}).
\end{itemize}

Our experiments suggest that adaptive thresholding could overcome these limitations. The implementation described in Section~\ref{sec:safeplan} successfully balances exploration and safety, as demonstrated in Figure~\ref{fig:cceg}~and~\ref{fig:ccel}.

\subsection{Switching Strategy}

We have discussed how using a world model enables the construction of local plans, which can yield more accurate behavior over short horizons. However, for long-term predictions, this approach may suffer from uncertainty and computational inefficiency (plan computation time extremely growths with the horizon). In SPOWL, we address this by using short horizons ($H = 3$) while combining plan suggestions with policy actions to select the optimal decision.

We evaluate three decision-making variants in the environment.
\begin{itemize}
    \item \textbf{Policy-only:} relying solely on the learned policy. 
    \item \textbf{Plan-only:} using only plans generated from each state.
    \item \textbf{Dynamic switching:} alternating between policy and plans during execution.
\end{itemize} 

\begin{figure}[h]
    \centering
    \includegraphics[width=0.9\columnwidth]{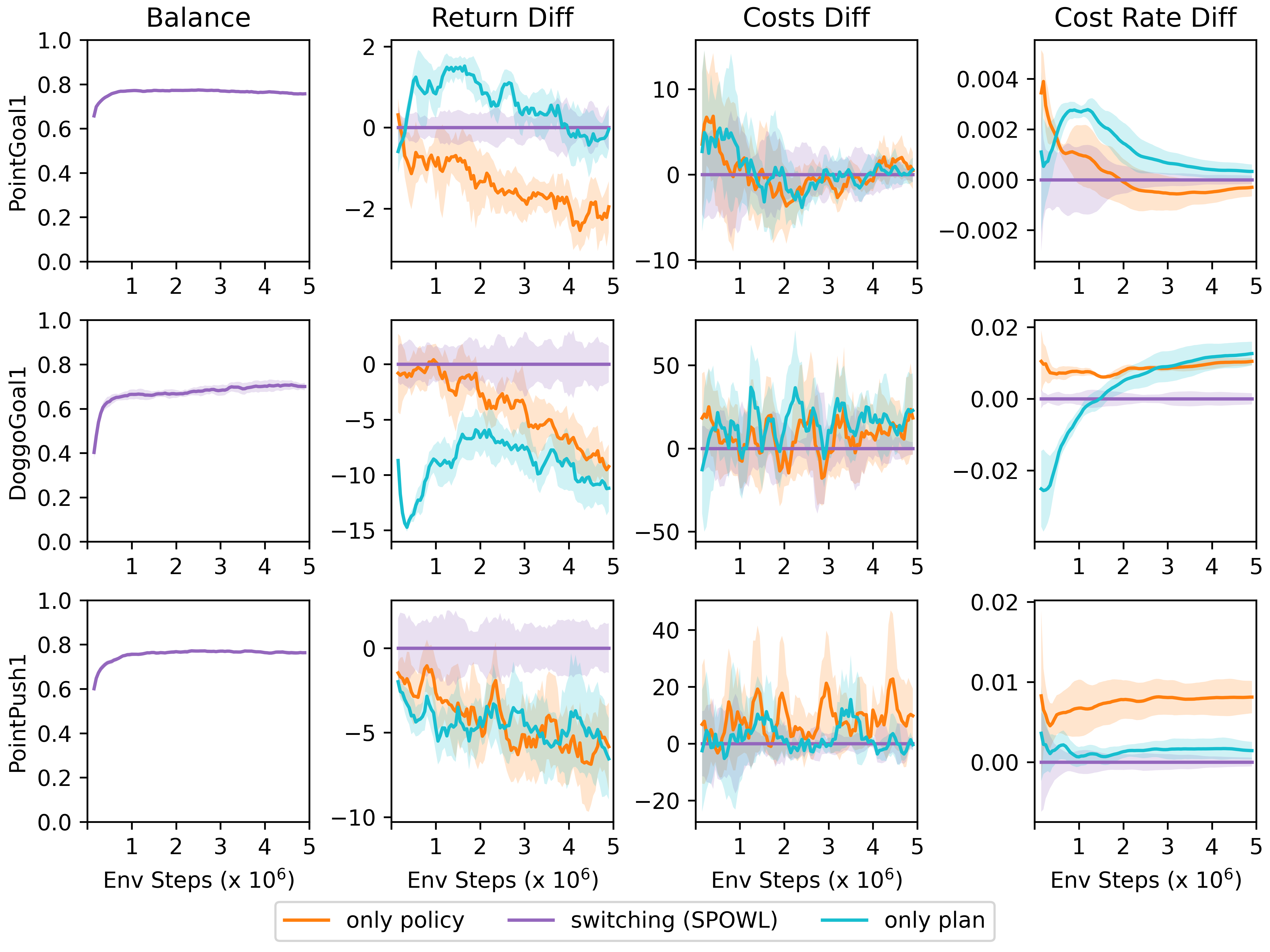}
    \caption{Performance comparison of decision-making approaches relative to SPOWL. Metrics with ``Diff'' show differences from SPOWL's performance (approach minus SPOWL): Return Diff, Costs Diff, and Cost Rate Diff. Balance represents the fraction of planning-based actions in SPOWL (0: pure policy, 1: pure planning).}
    \label{fig:switch}
\end{figure}

These variants were tested across tasks of varying complexity and type (see Figure~\ref{fig:switch}).

\textbf{Planning} demonstrates advantages in tasks with simple dynamics, where the world model can be learned quickly (Return Diff in Figure~\ref{fig:switch}, PointGoal1). However, the high imbalance between safe and unsafe transitions leads to slower learning of cost components (Cost Rate Diff in Figure~\ref{fig:switch}, PointGoal1). Planning is particularly effective in tasks requiring manipulation of external objects, as future predictions help avoid hazardous scenarios caused by poor object placement (Cost Rate Diff in Figure~\ref{fig:switch}, PointPush1).

\textbf{The policy variant} shows superior performance in more complex tasks (DoggoGoal1), exhibiting rapid performance growth during training. The plan-based variant only matches this performance near convergence (Return Diff in Figure~\ref{fig:switch}, DoggoGoal1). The initially lower cost rate for planning results from the challenges in precise body control during early learning phases (can not move in a direction).

\textbf{SPOWL's dynamic switching} strategy consistently outperforms using either policies or plans independently. Analysis of the Balance metric\footnote{The amount of actions come frome planning} in Figure~\ref{fig:switch} reveals that while planning contributes over 60\% of actions (demonstrating their general reliability), the balanced distribution between planning and policy actions confirms both components are necessary for optimal performance.

Our experimental results (see Figure~\ref{fig:switch}) show that SPOWL achieves superior rewards across most tasks. The synergistic combination of planning and policy in SPOWL is particularly evident in the PointPush1 results (Figure~\ref{fig:switch}), where it significantly outperforms either approach used independently.

\subsection{Ablation Study of Design Choices}
\label{subsec:ablate}

\begin{table*}[]
\caption{Baselines comparison. TD-MPC2 is only one that does not consider safety. (2) - denotes cost threshold for the episode is equal 2, the same for (25), but it equals 25. In tasks abbreviation P, C, D, A ---denotes Point Car, Doggo and Ant agent bodies. While G1, G2, P1, B1 are the type of task: Goal1, Goal2, Push1, Button1 \citep{ji2023safetygymnasium}. Metrics $R$---episode reward, $C$---episode cost violation, $\rho$---cost rate. All metrics provided after 5 million enviroment interaction steps. For learning curves see Supplementary materials.}\label{tab:sg6a}
\centering
\begin{tabular}{lc|llll|lll} 
\toprule
Task & Metric & \textbf{SPOWL} & TD-MPC2 & SafeDreamer & PPOLag(2) & PPOLag(25) & CPO(25) & CVPO(25)\\
\midrule
\multirow{3}{*}{PG1}
    & $R$ & $\mathbf{26.9 \pm 0.4}$  & $27.5 \pm 0.2$ & $12.0\pm4.0$ & $8.5\pm3.5$ & $14.5\pm0.6$ & $19.2\pm3.0$ & $15.3\pm6.4$\\
    & $C$ & $\mathbf{1.5 \pm 1.1}$ & $52.3\pm8.6$ & $10.3\pm9.5$ & $13.5\pm7.2$ & $19.7\pm3$ & $31.0\pm5.8$ & $22.9\pm8.5$\\
    & $\rho \times 10^4$ & $\mathbf{133 \pm2}$ & $515\pm9$ & $160\pm16$ & $222\pm19$ & $361\pm20$ & $275\pm21$ & $296\pm18$\\
\midrule
\multirow{3}{*}{PG2}
    & $R$ & $\mathbf{13.7 \pm 3.2}$ & $27.4\pm0.5$ & $5.2\pm3.9$ & $1.8\pm1.7$ & $2.0\pm0.9$ & $3.2\pm3.2$ & $2.0\pm1.2$\\
    & $C$ & $4.8 \pm 1.2$ & $179.0\pm12.2$ & $\mathbf{0\pm0}$ & $35.9\pm14.3$ & $33.7\pm16.3$ & $26.1\pm0.8$ & $22.6\pm21.3$\\
    & $\rho \times 10^4$ & $280\pm5$ & $1641\pm18$ & $\mathbf{221\pm32}$ & $397\pm53$ & $570\pm73$ & $354\pm45$ & $391\pm6$\\
\midrule
\multirow{3}{*}{PP1}
    & $R$ & $\mathbf{18.7\pm1.4}$ & $20.8\pm0.6$ & $12.5\pm7.2$ & $1.5\pm0.6$ & $4.4\pm2.7$ & $2.4\pm1.8$ & $10.4\pm2.7$\\
    & $C$ & $\mathbf{1.8\pm2.5}$ & $34.2\pm11.6$ & $7.3\pm10.4$ & $33.5\pm16.4$ & $24.2\pm12.4$ & $25.5\pm8.1$ & $25.8\pm9.8$\\
    & $\rho \times 10^4$ & $103\pm5$ & $353\pm6$ & $\mathbf{71\pm3}$ & $123\pm13$ & $276\pm16$ & $254\pm7$ & $253\pm12$\\
\midrule
\multirow{3}{*}{PB1} 
    & $R$ & $\mathbf{10.0\pm1.5}$ & $33.4\pm0.4$ & $0.1\pm1.4$ & $0.2\pm0.4$ & $3.7\pm2.5$ & $1.3\pm0.6$ & $\mathbf{10.1\pm1.7}$\\
    & $C$ & $5.4\pm2.7$ & $103.3\pm3.1$ & $\mathbf{0.3\pm0.5}$ & $43.7\pm35.4$ & $33.6\pm7.7$ & $22.8\pm5.4$ & $53.6\pm26.5$\\
    & $\rho \times 10^4$ & $269\pm3$ & $1219\pm10$ & $\mathbf{159\pm3}$ & $395\pm142$ & $392\pm27$ & $336\pm12$ & $358\pm19$\\
\midrule
\multirow{3}{*}{CG1}
    & $R$ & $\mathbf{31.3\pm0.8}$ & $34.8\pm0.3$ & $6.6\pm1.2$ & $3.5\pm3.2$ & $18.4\pm3.9$ & $21.6\pm2.4$ & $20.2\pm2.1$\\
    & $C$ & $1.0\pm1.0$ & $51.2\pm7.9$ & $\mathbf{0\pm0}$ & $7.5\pm6.8$ & $30.5\pm5.0$ & $25.6\pm2.8$ & $34.8\pm3.4$\\
    & $\rho \times 10^4$ & $\mathbf{113\pm2}$ & $563\pm11$ & $\mathbf{127\pm15}$ & $214\pm6$ & $345\pm9$ & $321\pm10$ & $283\pm10$\\
\midrule
\multirow{3}{*}{DG1}
    & $R$ & $\mathbf{19.0\pm1.9}$ & $27.9\pm1.5$ & $1.1\pm0.9$ & $0\pm0$ & $0\pm0$ & $0\pm0$ & $0\pm0$\\
    & $C$ & $0.3\pm0.4$ & $72.6\pm20.6$ & $0\pm0$ & $0\pm0$ & $21.8\pm30.8$ & $0\pm0$ & $0\pm0$\\
    & $\rho \times 10^4$ & $286\pm15$ & $659\pm33$  & $260\pm70$ & $249\pm15$ & $376\pm70$ & $224\pm79 $ & $332\pm24$\\
\midrule
\multirow{3}{*}{AG1}
    & $R$ & $\mathbf{38.9\pm1.4}$ & $61.0\pm2.6$ & $0.2\pm0.2$ & $0.2\pm0.2$ & $5.1\pm3.9$ & $0\pm0$ & $4.3\pm1.9$\\
    & $C$ & $2.0\pm0.2$ & $70.9\pm6.0$ & $0\pm0$ & $0\pm0$ & $23.6\pm18.2$ & $0\pm0$ & $41.5\pm44.1$\\
    & $\rho \times 10^4$ & $275\pm6$ & $682\pm6$ & $218\pm58$ & $220\pm141$& $638\pm115$ & $466\pm38$ & $598\pm33$\\
\bottomrule
\end{tabular}
\end{table*}

Cost components (see Figure~\ref{fig:architecture}) are crucial for safe agent behavior. Following TD-MPC2 \citep{hansen2024tdmpc2}, we implement similar ensemble-based components with subsampled aggregation, and conduct detailed analysis of sizes and aggregation methods to examine their impact on both safety and reward performance.

\begin{figure}[h]
    \centering
    \includegraphics[width=0.8\columnwidth]{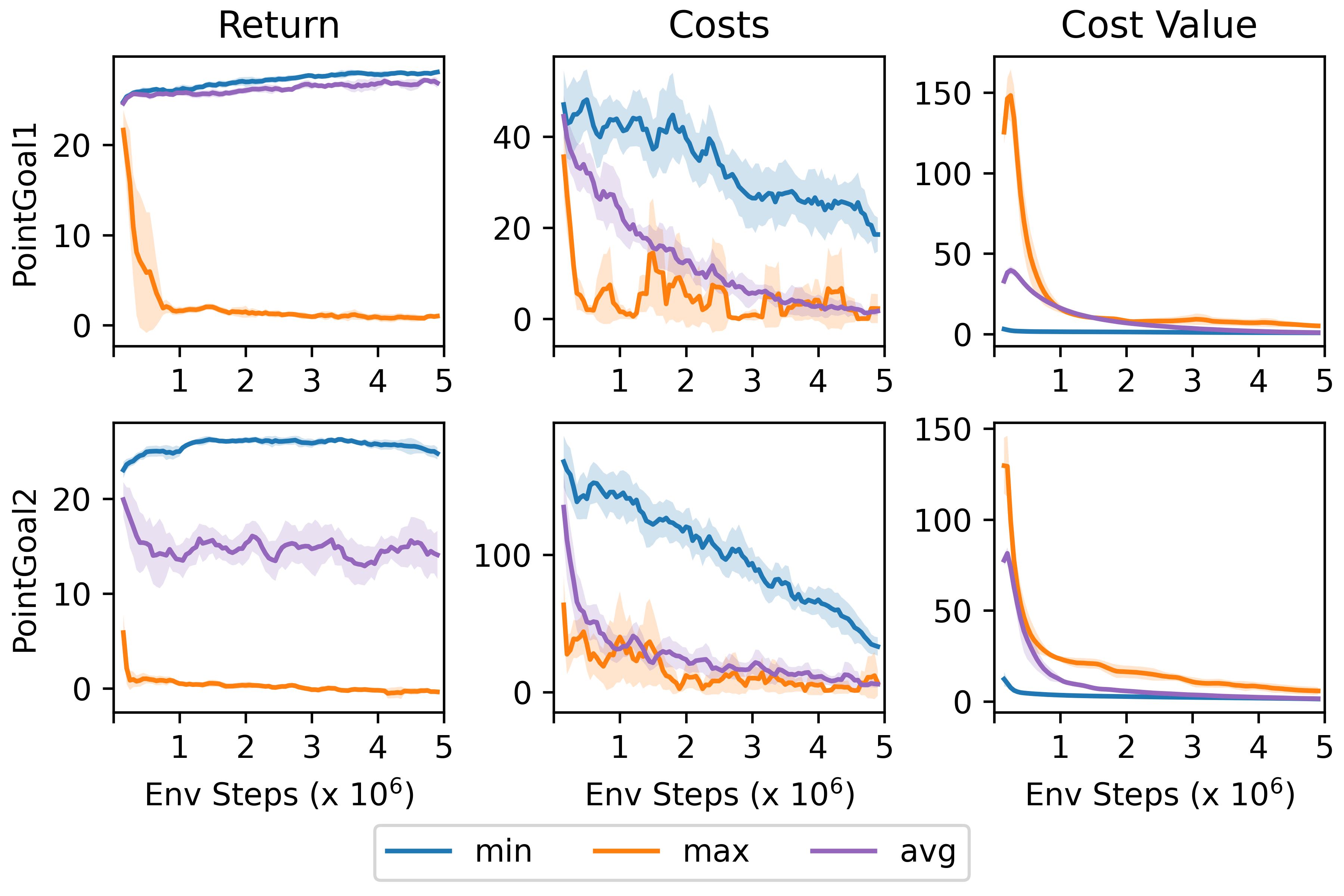}
    \caption{Comparison of aggregation methods for target cost estimation. Average avoids both overestimation and underestimation extremes.}
    \label{fig:cvaltarg}
\end{figure}

For target evaluation (Equation~(\ref{eq:qtargets})), we compare three aggregation methods: minimum, maximum, and average over the ensemble. The results (Figure~\ref{fig:cvaltarg}) demonstrate that maximum aggregation leads to overestimation, degrading solution quality, while minimum aggregation causes underestimation that results in dangerous cost violations. The average method provides the most balanced estimation without these extremes.

\begin{figure}[h]
    \centering
    \includegraphics[width=\columnwidth]{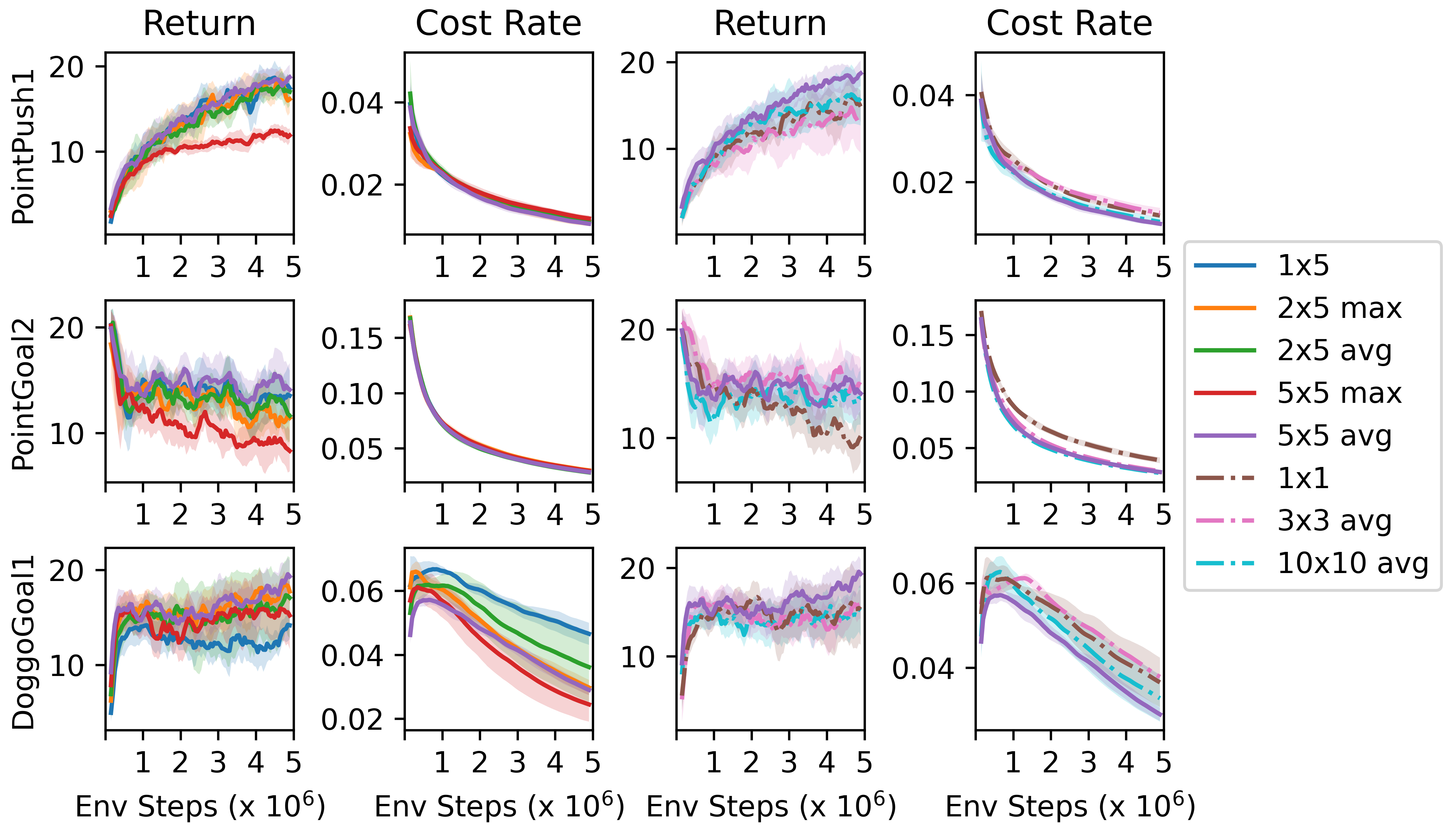}
    \caption{Ensemble ablation study. Notation XxY indicates subsample size X and ensemble size Y.}
    \label{fig:ens}
\end{figure}

For the Augmented Lagrangian policy loss ($\Delta$ in Equation~(\ref{eq:lag})), we analyze subsampling effects across ensemble sizes, subsample counts, and aggregation methods (focusing on average and maximum, excluding minimum due to its safety risks). Experiments (Figure~\ref{fig:ens}) reveal that larger subsamples improve safety by reducing cost violations, and maximum aggregation provides greater safety than average. However, in tasks like PointGoal2 and PointPush1, maximum aggregation degrades reward performance without corresponding safety gains. The right portion of Figure~\ref{fig:ens} further reveals that single-head ensembles underperform while oversized ensembles do not yield any significant improvements. Based on these findings, we use the 5x5 average scheme for SPOWL as it provides sufficiently tight cost estimation without the degradation in performance.

\begin{figure}[h]
    \centering
    \includegraphics[width=0.9\columnwidth]{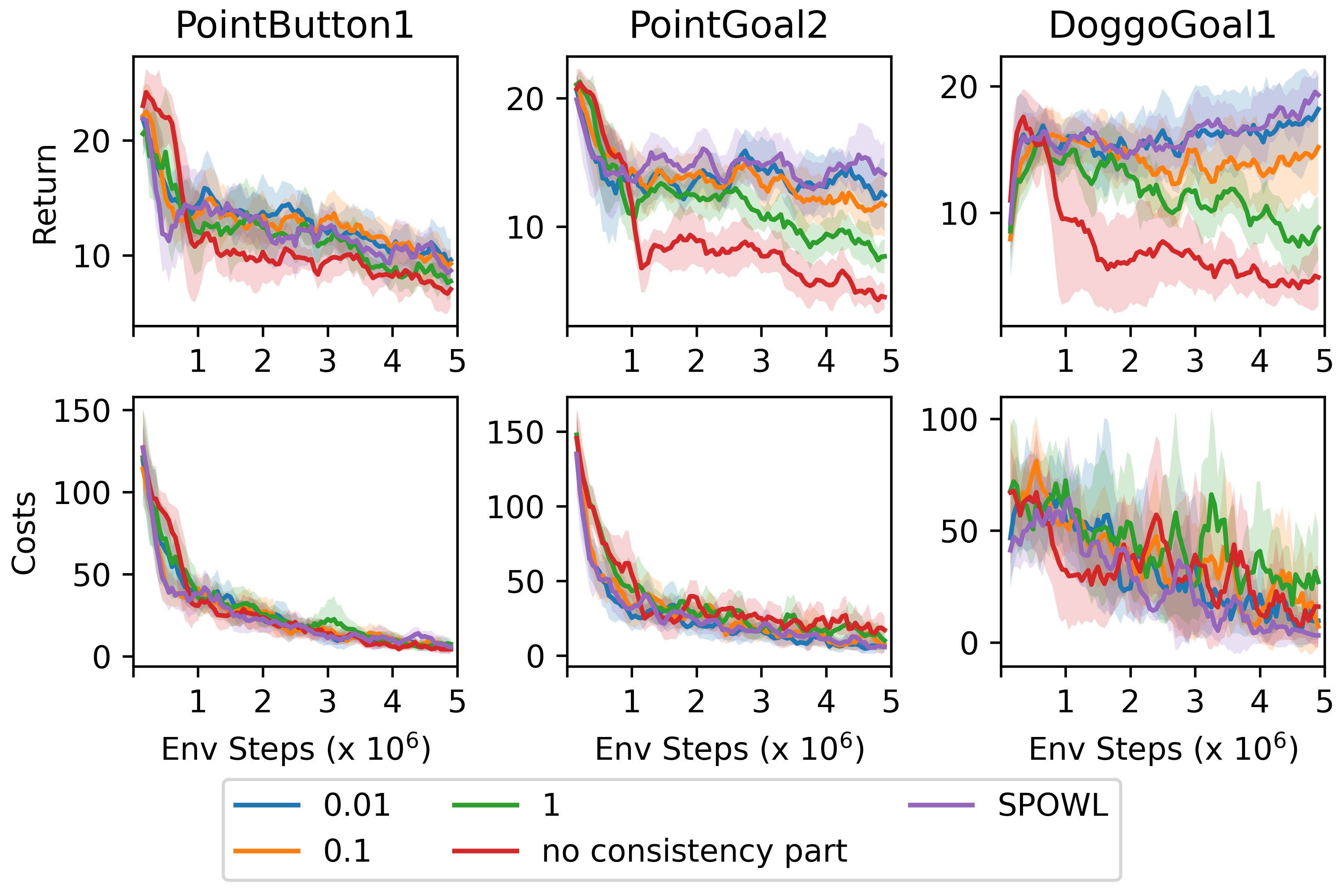}
    \caption{Decoder ablation study with loss weights 0.01, 0.1, and 1. "No consistency" indicates weight=0 for consistency loss while keeping decoder weight at 20 (usual for consistency loss).}
    \label{fig:decoder}
\end{figure}

The implicit world model in SPOWL shows advantages over traditional encoder-decoder architectures, consistent with \citet{hansen2024tdmpc2}. When we modify SPOWL to include an auxiliary decoder (used only for additional loss computation), the results (Figure~\ref{fig:decoder}) show no safety improvements but significant degradation in reward performance, suggesting this component interferes with value optimization.

\subsection{Comparative Analysis}

We evaluate several Safe RL algorithms to provide a comprehensive comparison. For fairness, all algorithms use vector observations (proprioception and LiDAR data). The default baseline for unsafe behavior is TD-MPC2 \citep{hansen2024tdmpc2}.

The closest to our approach is SafeDreamer---a model-based algorithm designed to achieve near-zero cost violations. Our experimental results (see Table~\ref{tab:sg6a}) show that SafeDreamer attains very low safety violations, but this comes at a significant cost to reward performance. In some tasks (Doggo and Ant Goal1), SafeDreamer fails to learn entirely. For AntGoal, SafeDreamer cannot even begin collecting data because it requires episodes longer than 16 steps, while the task may terminate early.

We also compare traditional model-free methods: PPOLag \citep{stooke2020pidlag} and CPO \citep{stooke2020pidlag}, both on-policy algorithms. We demonstrate that simply lowering the cost threshold is insufficient for adequate performance (see Table~\ref{tab:sg6a}). PPOLag(2) (cost limit = 2) performs the worst in both reward and safety, while PPOLag(25) (cost limit = 25) achieves significantly better results.

CVPO, a model-free off-policy algorithm \citep{liu2022cvpo}, outperforms on-policy variants (all algorithms are limited to 5 million training steps). It even matches SPOWL’s performance on PB1, though at the expense of safety. Notably, model-free algorithms successfully learn in complex environments (AntGoal1, DoggoGoal1), whereas SafeDreamer stagnates. In contrast, SPOWL delivers dominant reward performance while maintaining near-zero cost violations.

\section{Conclusion}

This paper presents SPOWL, a novel framework that effectively combines policy optimization with model-based planning for safe reinforcement learning. Thanks to our carefully designed framework, \textbf{SPOWL outperforms existing methods and delivers robust performance across a variety of tasks with different dynamics and embodiments}. We provide a stable solution for challenging continuous control tasks, including challenging high-dimensional embodiments (e.g., Ant and Doggo).

The current implementation has several limitations that suggest important directions for future research. The planning component, while efficient due to our JAX \citep{jax2018github} implementation, still requires more computation than purely model-free approaches. Furthermore, the framework currently processes only vector observations, leaving extension to pixel-based inputs as valuable future work. Most notably, while SPOWL demonstrates excellent final performance and safety, the initial training phases may experience safety violations before optimal policies emerge. This observation motivates our ongoing work to develop a framework that preserve SPOWL's performance advantages while further reducing early-training safety risks. Such improvements would significantly enhance the framework's applicability to real-world scenarios where strict safety constraints must be maintained throughout the entire learning process.






\bibliography{mybibfile}

\newpage
\appendix
\section{Algorithms}
\begin{algorithm}[h]
   \caption{SelectElites}\label{algo:selectelites}
   \begin{algorithmic}[]
       \Require { Action sequences $\mathcal{A}$, Policy prior $\mathcal{A}_\pi$, Thresholds $d^R$, $d^c$, Value estimates for every action sequence $J^M,J^M_c$.}
       \State $\mathcal{A}_\text{impr} \gets \{\mathbf{a} \in \mathcal{A} \mid J^M(\mathbf{a}) \geq d^R \wedge J^M_c(\mathbf{a}) \leq d^c\}$
        \If{$\mathcal{A}_\text{impr}=\emptyset$}
            \State $\mathcal{A}_\text{elites} \gets \mathcal{A}_\pi$
        \ElsIf{$|\mathcal{A}_\text{impr}| \leq k$}
            \State $\mathcal{A}_\text{elites} \gets \mathcal{A}_\text{impr}$
        \Else
            \State $\mathcal{A}_\text{elites} \gets \operatorname{top}_k(\mathcal{A}_\text{impr}, J^M)$
        \EndIf
        \State \textbf{return} $\mathcal{A}_\text{elites}$ 
   \end{algorithmic}
\end{algorithm}

\begin{algorithm}[h]
\caption{SPOWL Model Predictive Control}\label{alg:adaptiveplan}
\begin{algorithmic}[]
\Require Environment ENV, safe policy $\pi$, world model $M$ with ensemble value functions $\{\hat{Q}_j, Q_{c,j}\}$, episode length $T$
\State $s_0 \gets \text{ENV.reset()}$
\State $\mathbf{a}_0 \gets \mathbf{0}$ \Comment{Initialize action sequence}
\For{$t \gets 0$ to $T-1$}
    \State $z_t \gets \text{Encoder}(s_t)$ \Comment{Encode to latent space}
    \State $\mathcal{A}_\pi \gets \text{GeneratePriorSet}(z_t,\pi,M)$
    \State $\bm{\sigma}_0 \gets (\sigma_{\text{init}}, \dots, \sigma_{\text{init}})$
    \State $\bm{\mu}_0 \gets \mathbf{a}_t=(a_t, a_{t+1}, \dots, a_{t+H-1})$
    
    \For{$i \gets 0$ to $N-1$} \Comment{Planning iterations}
        \State $\mathcal{A} \gets \mathcal{A}_i \cup \mathcal{A}_\pi$, $\mathcal{A}_i \sim \mathcal{N}(\bm{\mu}_i, \bm{\sigma}_i)$
        \State $J^M, J^M_c \gets \text{EstimateValue}(z_t,\mathcal{A},\pi,M)$ 
        \State $d^R, d^c \gets \text{SetThresholds}(J^M, J^M_c, \mathcal{A}, \mathcal{A}_\pi)$ 
        \State $\mathcal{A}_\text{elites} \gets \text{SelectElites}(d^R, d^c, \mathcal{A}, \mathcal{A}_\pi, J^M, J^M_c)$
        \State $\bm{\mu}_{i+1}, \bm{\sigma}_{i+1} \gets \text{mean}(\mathcal{A}_\text{elites}), \text{var}(\mathcal{A}_\text{elites})$
    \EndFor
    
    \State $a_\text{plan}=a_t\leftarrow\mathbf(a_t, a_{t+1}, \dots, a_{t+H-1}) \sim \mathcal{A}_\text{elites}$
    \If{$\hat{Q}_\text{avg}(z_t,a_\text{plan}) \geq \hat{Q}_\text{avg}(z_t,\pi(z_t))$ and \State $\hat{Q}_{c,\text{avg}}(z_t,a_\text{plan}) \leq \hat{Q}_{c,\text{avg}}(z_t,\pi(z_t))$}
        \State $a_\text{env} \gets a_\text{plan}$
    \Else
        \State $a_\text{env} \gets \pi(z_t)$
    \EndIf
    
    \State $s_{t+1}, r_{t+1}, c_{t+1}, d_{t+1} \gets \text{ENV.step}(a_\text{env})$
    \If{$d_{t+1}$ is True} \Comment{Episode termination}
        \State \textbf{break}
    \EndIf
    \State $\mathbf{a}_{t+1}=(a^N_{t+1}, \dots, a^N_{t+H-1}, 0) \leftarrow \bm{\mu}_N$ \Comment{Shift sequence}
\EndFor
\end{algorithmic}
\end{algorithm}

\section{SG6 Set of Tasks}
\begin{figure}[h]
    \centering
    \includegraphics[width=0.9\columnwidth]{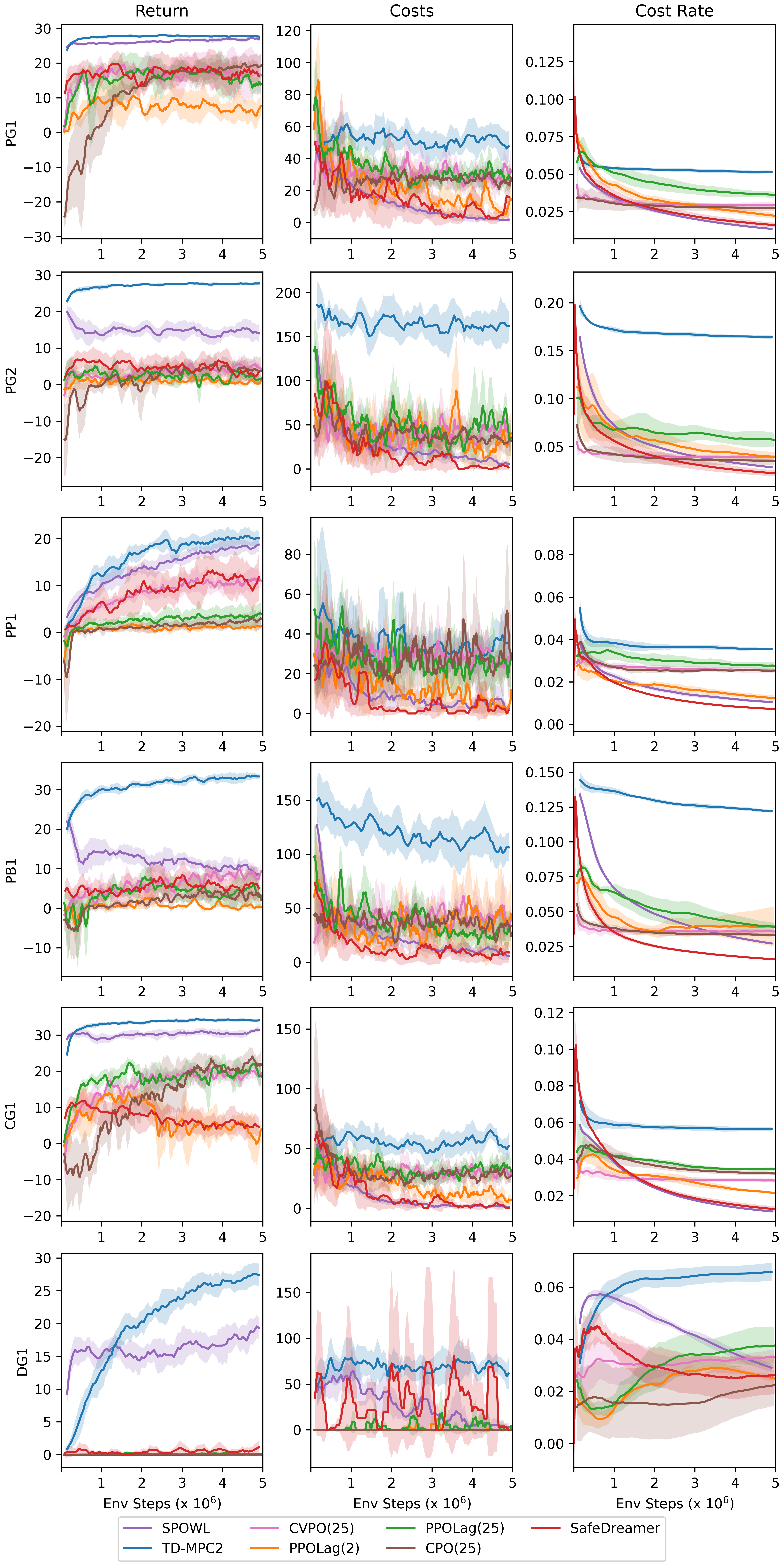}
    \caption{Learning curves for the main evaluation metrics. Abbreviations for tasks are PG1-PointGoal1, PG1-CarGoal1, DG1-DoggoGoal1, PG2-PointGoal2, PP1-PointPush1, PB1-PointButton1.}
\end{figure}

\end{document}